%% file: main.tex
\documentclass[conference]{IEEEtran}
\IEEEoverridecommandlockouts

\usepackage{cite}
\usepackage{amsmath,amssymb,amsfonts}
\usepackage{algorithmic}
\usepackage{graphicx}
\usepackage{textcomp}
\usepackage{xcolor}
\usepackage[hidelinks]{hyperref}
\usepackage{threeparttable}

\def\BibTeX{{\rm B\kern-.05em{\sc i\kern-.025em b}\kern-.08em
    T\kern-.1667em\lower.7ex\hbox{E}\kern-.125emX}}
\begin{document}

\title{Advancing Mental Disorder Detection: A Comparative Evaluation of Transformer and LSTM Architectures on Social Media}

\author{
    \IEEEauthorblockN{Khalid Hasan}
    \IEEEauthorblockA{
        \textit{Dept. of Computer Science} \\
        \textit{Missouri State University}\\
        Springfield, MO, USA \\
        kh597s@missouristate.edu
    }
    \and
    \IEEEauthorblockN{Jamil Saquer}
    \IEEEauthorblockA{
        \textit{Dept. of Computer Science} \\
        \textit{Missouri State University}\\
        Springfield, MO, USA \\
        jamilsaquer@missouristate.edu
    }
    \and
    \IEEEauthorblockN{Mukulika Ghosh}
    \IEEEauthorblockA{
        \textit{Dept. of Computer Science} \\
        \textit{Missouri State University}\\
        Springfield, MO, USA \\
        mghosh@missouristate.edu
    }
 }

\maketitle

\input{abstract}


\input{sections/introduction}
\input{sections/related_work}
\input{sections/problem_definition}
\input{sections/methodology}

\input{sections/results}

\input{sections/conclusion}

\bibliographystyle{IEEEtran}
\bibliography{references}

\end{document}

%% file: abstract.tex
\begin{abstract}

The rising prevalence of mental health disorders necessitates the development of robust, automated tools for early detection and monitoring. Recent advances in Natural Language Processing (NLP), particularly transformer-based architectures, have demonstrated significant potential in text analysis. This study provides a comprehensive evaluation of state-of-the-art transformer models (BERT, RoBERTa, DistilBERT, ALBERT, and ELECTRA) against Long Short-Term Memory (LSTM) based approaches using different text embedding techniques for mental health disorder classification on Reddit. We construct a large annotated dataset, validating its reliability through statistical judgmental analysis and topic modeling. Experimental results demonstrate the superior performance of transformer models over traditional deep-learning approaches. RoBERTa achieved the highest classification performance, with a 99.54\% F1 score on the hold-out test set and a 96.05\% F1 score on the external test set. Notably, LSTM models augmented with BERT embeddings proved highly competitive, achieving F1 scores exceeding 94\% on the external dataset while requiring significantly fewer computational resources. These findings highlight the effectiveness of transformer-based models for real-time, scalable mental health monitoring. We discuss the implications for clinical applications and digital mental health interventions, offering insights into the capabilities and limitations of state-of-the-art NLP methodologies in mental disorder detection.

\end{abstract}

%% file: sections/introduction.tex
\section{Introduction}

Mental health conditions such as depression, anxiety, and schizophrenia remain prevalent global health challenges. According to the World Health Organization (WHO), one in eight people will experience a mental illness during their lifetime~\cite{WorldHealth}, making early detection and intervention critical. Untreated mental health conditions often lead to serious functional impairment, reduced quality of life, and increased mortality risk~\cite{momen2022mortality}. These disorders' significant societal and economic impact underscores the need for effective monitoring and treatment tools~\cite{arias2022}.

Social media platforms have transformed research into mental health disorders, with sites like Reddit becoming spaces where individuals openly discuss their mental health experiences. Several subreddits are dedicated to mental health support where people share their thoughts, emotions, and experiences. Unlike traditional methods, which rely on surveys or clinical records, social media networks generate continuous, authentic mental health expressions in real time~\cite {sadagheyani2021investigating}. While analysis of this data offers valuable insights for developing early intervention tools, mining mental health content from social media faces distinct challenges, including linguistic variability, colloquial expressions, and unstructured text formats.

Most previous work related to mental health detection has applied traditional machine learning methods, such as Support Vector Machines, Naive Bayes, and Random Forests~\cite{iyortsuun2023review}. Although these methods have succeeded, they are still moderated; the intrinsic nature of these approaches simply cannot keep pace with the complexity of evolving natural language~\cite{karamizadeh2014advantage}. Discussions about mental health often rely on subtle cues, metaphorical language, and emotional undertones, making manual feature extraction inherently challenging. Therefore, most of these methods struggle to generalize across different datasets and contexts.

The emergence of deep learning has significantly advanced Natural Language Processing (NLP), enabling more nuanced feature representation directly from textual data. Recurrent Neural Networks (RNNs) and Long Short-Term Memory (LSTM) networks demonstrated early promise in sequence modeling tasks like text classification~\cite{kang2021classification}. However, these architectures struggle with capturing long-range contextual dependencies, which are particularly critical in complex domains such as mental health narrative analysis~\cite{wu2020review}. Conversely, transformer models represent a pivotal breakthrough, addressing many prior limitations through self-attention mechanisms. Models like Bidirectional Encoder Representations from Transformers (BERT) leverage sophisticated contextual understanding, achieving remarkable performance across diverse NLP tasks, from sentiment analysis to question answering and text classification~\cite{tunstall2022natural}. Although BERT-based approaches have outperformed traditional machine learning models, recent research shows they face limitations when processing clinical documents~\cite{gao2021limitations}. This motivated us to conduct research that investigates mental health disorders in a large dataset collected from Reddit by utilizing state-of-the-art deep-learning techniques. 

Our work investigates the performance of transformer-based and LSTM-based architectures in detecting mental health posts from Reddit posts. Unlike prior research, our study does not focus on a certain type of mental health disease, such as depression or suicidal ideation. The specific task we tackle is a binary classification of identifying posts indicative of any mental health disorder from posts written by individuals who do not have such conditions. This paper primarily presents a performance comparison among transformer models like BERT, RoBERTa, DistilBERT, ALBERT, ELECTRA, and LSTM-based models with advanced techniques such as attention mechanisms. After presenting the strengths and weaknesses of such models, practical insights can be gained in developing automated tools for mental health monitoring.

There are two primary objectives of our research: first, assessing the performance of deep learning transformer-based models in detecting posts with signs of mental health on social media, and second, comparing the performance of transformers with LSTM-based models. The methods designed to achieve these goals contribute to developing practical tools for mental health monitoring on social media platforms.

\textbf{The main contributions of the work in this paper are:}

\begin{enumerate}
    \item Development and annotation of a large dataset from Reddit containing posts of people with mental disorders and a control group. We make the dataset available to researchers upon request.
    
    \item Validation of annotation quality through statistical judgmental analysis and Latent Dirichlet Allocation (LDA) topic modeling of the mental health-related posts. 
    
    \item Performance evaluation of state-of-the-art transformer architectures, including BERT and RoBERTa, for mental health content classification using our annotated dataset.
    
    \item Comparative analysis of LSTM models across various text embedding approaches, such as BERT, GloVe, and Word2Vec, for mental health condition detection.
\end{enumerate}

%% file: sections/related_work.tex
\section{Related Work}

The computational analysis of mental health has attracted considerable attention in recent years, particularly with the growing use of social media for psychological expression~\cite{kim2021machine, coppersmith-etal-2015-adhd}. Early efforts to identify mental health conditions from text relied primarily on feature engineering and traditional machine learning techniques, such as Support Vector Machines (SVMs), Naive Bayes, and Random Forests~\cite{kowsari2019text, de2013predicting}. For instance, Tsugawa et al. achieved 69\% accuracy in detecting depression risk from Japanese text using an SVM classifier with handcrafted linguistic features~\cite{tsugawa2015recognizing}. Similarly, De Choudhury et al. employed an SVM classifier to analyze semantic and behavioral attributes from Twitter data, achieving 70\% accuracy in identifying individuals at risk of depression~\cite{de2013predicting}.

Although these traditional methods demonstrated promising results in specific contexts, their dependence on manual feature engineering posed limitations in generalizing across diverse linguistic expressions of mental health conditions~\cite{kowsari2019text}. While feature engineering showed promise in time series analysis~\cite{fahim2024serein}, the intricate nature of semantic and contextual cues in user-generated content necessitated more sophisticated learning frameworks capable of capturing deeper patterns. This led to the adoption of deep learning models, particularly Recurrent Neural Networks (RNNs) and Long Short-Term Memory (LSTM) networks, which excel at processing sequential data and extracting contextual dependencies~\cite{garg2023mental}. For example, Ameer et al.~\cite{ameer2022mental} utilized both LSTM and Bi-LSTM architectures to perform linguistic analysis on social media content for mental illness detection. In contrast, their F1-scores did not exceed 0.79 for either model. Despite the advancements offered by deep learning, these models still faced challenges in capturing nuanced semantics and contextual dependencies, particularly when identifying subtle expressions of mental distress with high precision~\cite{wu2020review}.

The advent of transformer-based models, particularly BERT, introduced by Devlin et al.~\cite{devlin2019bert}, has marked a significant breakthrough in natural language processing (NLP) tasks, including mental health classification. The self-attention mechanism in BERT enables the model to capture long-range dependencies more effectively than traditional RNNs or LSTMs, making it particularly suited for analyzing complex textual data. For instance, Hasan et al.~\cite{hasan2024comparative} leveraged BERT embeddings combined with a BiLSTM network to detect suicide-related content on social media, achieving an accuracy of above 92\%. This highlights the potential of transformer-based hybrid models in cross-domain mental health detection. Qasim et al.~\cite{qasim2022fine} fine-tuned BERT-based models across different Twitter datasets, revealing promising results in transfer learning for mental disorder detection.

Along with classification models, topic modeling techniques have been widely used to analyze mental health discussions. Latent Dirichlet Allocation (LDA) has been one of the most prominent methods in uncovering thematic structures in online conversations~\cite{resnik2015beyond,islam2021mining}. For example, Resnik et al.~\cite{resnik2015beyond} applied LDA to identify distress signals in messages from mental health forums, revealing major themes such as anxiety, suicidal ideation, and coping mechanisms. More recently, neural topic modeling approaches have sought to enhance LDA by incorporating word embeddings and deep learning components to improve topic coherence and interpretability~\cite{dieng2020topic}. These approaches provide further insights into the underlying concerns expressed in mental health-related posts, which can complement predictive models in understanding user behavior.

Dataset quality and annotation reliability remain crucial considerations in mental health analysis. Previous studies identified the importance of inter-annotator agreement metrics, such as Cohen's $\kappa$~\cite{Cohen1960} and Krippendorff's $\alpha$~\cite{krippendorff2004}, for the credibility of a labeled dataset~\cite{northcutt2021pervasive, chakraborty2024comparative}. Northcutt et al.~\cite{northcutt2021pervasive} demonstrated that inconsistent manual labeling can significantly impact model performance and lead to biased results. To address these challenges, recent studies have increasingly adopted robust annotation protocols that combine human expertise with automated mechanisms to enhance dataset validity. 

Building on these contemporary advancements, our work investigates the performance of transformer-based models for mental disorder detection and compares them with LSTM-based approaches. We make several key contributions to the field. First, we curate a large and representative dataset of mental health-related posts from Reddit, ensuring robust sampling and relevance. Using this dataset, we train and evaluate various transformer-based models, assessing their effectiveness in identifying content related to mental distress. Second, we propose two methodological extensions: (1) integrating topic modeling and judgmental analysis to enhance dataset reliability and reduce subjectivity in labeling, and (2) conducting a comparative analysis of different LSTM models using various text embeddings, benchmarking their performance against transformer architectures. By combining state-of-the-art NLP techniques with rigorous statistical validation, our study provides a comprehensive evaluation of advanced deep-learning methods for automated mental health monitoring. These insights contribute to developing more effective tools for early intervention and mental health support.

%% file: sections/problem_definition.tex
\section{Problem Definition}

The principal objective of this research is to classify social media posts from Reddit related to mental health disturbance using transformer-based and LSTM-based deep learning models. The task is formulated as a binary classification problem, aiming to identify whether a post is associated with a mental health disorder.

\textbf{Problem Statement} - \textit{
Given a labeled dataset D = \{$d_1$, $d_2$, ..., $d_n$\} of $n$ posts, develop text classification models trained to map each post $d_i$ to a label $p_i \in \{0, 1\}$, where $0$ indicates the absence of mental illness, while $1$ indicates its presence.
}
 
To address this problem, we formulate three research questions, each focused on evaluating the performance of classification models. The questions are as follows.

\begin{enumerate}
    \item How competitive are transformer-based models in identifying posts of mental illness?

    \item How does integrating different word embeddings, such as BERT, GloVe, and Word2Vec, affect the performance of LSTM-based models, and what contribution does the attention mechanism make toward their effectiveness?

    \item How robust are the models in terms of computational complexity?
\end{enumerate}

%% file: sections/methodology.tex
\section{Methodology}

\begin{figure}
    \centering
    \includegraphics[width=1\linewidth]{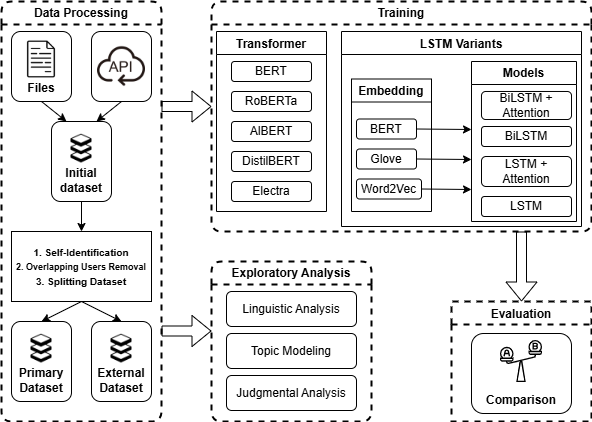}
    \caption{An Overview of our Research Architecture}
    \label{fig:research_architecture}
\end{figure}

Figure~\ref{fig:research_architecture} provides a comprehensive overview of our research framework, starting with data extraction and concluding with model comparison.

\input{sections/methodology/dataset}
\input{sections/methodology/exploratory_analysis}

\input{sections/methodology/classification_models}

%% file: sections/methodology/dataset.tex
\subsection{Dataset}

\subsubsection{Data Collection} 
We collected Reddit posts through December 2022 using the Pushshift\footnote{https://pushshift.io/} API to create a mental health discourse dataset. The data includes posts from mental health-focused subreddits (r/ADHD, r/anxiety, r/bipolar, r/depression, r/CPTSD, r/schizophrenia, r/BPD, r/SocialAnxiety, r/offmychest) and control group subreddits (r/confidence, r/politics, r/bicycletouring, r/sports, r/travel, r/geopolitics). Subreddits were selected based on active user engagement and sufficient text volume to enable balanced distribution between mental health and control classes, as detailed in the Subsection \ref{subsec:annotation}. Table~\ref{tab:mental_health_dataset} summarizes the initial collection of mental health condition posts.

\input{tables/mental_disorders_posts_count}

\input{sections/methodology/data_processing}

\input{tables/final_dataset}

\subsubsection{Annotation}
\label{subsec:annotation}
For the annotation of the dataset, posts were classified based on their subreddit of origin. Posts from mental health-related subreddits, such as r/ADHD, r/anxiety, r/bipolar, r/depression, r/CPTSD, r/schizophrenia, r/BPD, r/SocialAnxiety, and r/offmychest, were categorized as relating to mental disorders. These subreddits serve as support communities where members, often in distress, share their struggles and experiences with mental health, treatment, and symptoms of mental illness. It is reasonable to assume that posts from these subreddits reflect some form of mental disturbance. Many posts express emotional distress, risk behaviors, suicidal ideation, or experiences with therapy and medication, which makes them highly relevant to our classification task.

In contrast, posts from general discussion subreddits, such as r/politics, r/travel, r/sports, r/geopolitics, and r/bicycletouring, were classified into the control group. These subreddits focus primarily on broad societal topics unrelated to mental health, making them an appropriate neutral comparison group. Table~\ref{tab:final_dataset_overview} provides an overview of the annotated dataset.

To further ensure the veracity of labels within our dataset, we imposed two extra verification stages on the above methodology: \textbf{Topic Modeling} and \textbf{Judgmental Analysis} as discussed in subsections~\ref{subsubsec:topic_modeling} and ~\ref{subsubsec:judgmental_analysis}.

By applying heuristics for self-identification, filtering by user-based methods, and incorporating additional validation steps, we have created a highly reliable and representative dataset for training text classification models in the domain of mental disorder detection.

%% file: tables/mental_disorders_posts_count.tex
\begin{table}[htb]
    \centering
    \caption{Dataset Statistics: Initial and Processed Counts of Posts for Mental Health Conditions}
    \begin{tabular}{l|r|r}
        \hline
        \textbf{Subreddit} & \textbf{Initial Count (K)} & \textbf{Processed Count (K)} \\
        \hline
        ADHD & 643K & 254K \\
        Anxiety & 571K & 126K \\
        Bipolar & 303K & 111K \\
        CPTSD & 203K & 33K \\
        Depression & 1,516K & 222K \\
        Schizophrenia & 96K & 28K \\
        \hline
        Social Anxiety & 184K & 12K \\
        BPD & 295K & 80K \\
        Off My Chest & 1,607K & 33K \\
        \hline
        \textbf{Total} & \textbf{5,418K} & \textbf{899K} \\
        \hline
    \end{tabular}
    \label{tab:mental_health_dataset}
\end{table}

%% file: sections/methodology/data_processing.tex
\subsubsection{Data Processing}
\label{subsection-DataProcessing}

We followed the self-identification techniques previously used in analyzing mental health discourse~\cite{coppersmith-etal-2015-adhd, mitchell-etal-2015-quantifying, kim2023understanding}. We crafted specific regular expressions to capture users who explicitly stated their mental health condition. Moreover, we identified users who had posted in both mental health and control group subreddits to prevent contamination between groups. Then, we followed Cohan et al.~\cite{cohan-etal-2018-smhd} in removing posts from these overlapping users to create a clear separation between the two categories. The number of posts collected initially and those remaining after preprocessing are reported in Table~\ref{tab:mental_health_dataset}.

We then divided the entire dataset into two subsets: \textbf{primary} and \textbf{external}. The primary dataset consists of the subreddits: r/ADHD, r/anxiety, r/bipolar, r/depression, r/CPTSD, r/schizophrenia, r/confidence, r/politics, r/bicycletouring, and r/sports. This subset is used for model training, validation, and hold-out testing. The external dataset includes the posts from r/BPD, r/SocialAnxiety, r/offmychest, r/travel, and r/geopolitics, which do not overlap with the primary subset. This external dataset is used solely for external validation to assess the model's ability to generalize to unseen mental health conditions and control groups. We selected 144,000 labeled posts, with 120,000 allocated to the primary dataset and 24,000 to the external dataset. We applied stratified sampling to ensure a balanced dataset by selecting an equal number of posts from each subreddit community. Table~\ref{tab:final_dataset_overview} provides an overview of the final dataset composition.

Next, we performed a structured, step-by-step preprocessing of the raw text data to prepare it for analysis. The preprocessing steps included tokenization, cleaning, normalization, and lemmatization. First, we applied whitespace tokenization, which splits the text into individual words. We then used regular expressions during normalization to filter out irrelevant elements such as emojis, URLs, special characters, and digits, allowing us to retain only meaningful linguistic features. Finally, lemmatization was performed to reduce words to their base forms, enhancing consistency and generalizability in the textual representations.

For various analytical tasks, we utilized text data at different preprocessing stages. For instance, in part-of-speech (POS) tagging, we employed only tokenized text to maintain the original linguistic structure, which is essential for accurate tagging. In contrast, topic modeling was carried out using the fully processed, lemmatized dataset to enhance topic coherence. Likewise, for extracting mental health-related phrases, we worked with the original unprocessed text to capture contextual nuances and identify significant multi-word expressions.

This structured preprocessing approach ensured consistency across the dataset while optimizing it for various linguistic analyses, with minimal loss of critical contextual and structural information.

%% file: tables/final_dataset.tex
\begin{table}[htb]
    \caption{An overview of the Final Dataset}
    \centering
    \begin{tabular}{c|c|c|c}
        \hline
         & \textbf{Mental Disorders} & \textbf{Control Group} & \textbf{Total} \\
         \hline
        \textbf{Primary} & 60,000 & 60,000 & 120,000 \\
        \hline
        \textbf{External} & 12,000 & 12,000 & 24,000 \\
    \hline
    \end{tabular}
    \label{tab:final_dataset_overview}
\end{table}

%% file: sections/methodology/exploratory_analysis.tex
\subsection{Exploratory Analysis}

We have divided the analysis of our dataset into three stages to ensure a comprehensive understanding of the data. First, we perform a textual analysis for the posts under the category of mental disorder and control separately to discover some crucial linguistic patterns and insights. We subsequently apply topic modeling in the mental illness group to understand the nature of discussions by highlighting prevalent themes and psychological distress expressed in user-generated content. Finally, we carry out judgmental analysis over the whole dataset, systematically considering the quality and reliability of the annotations to ensure that the labels are valid for further modeling and analysis.

\input{sections/methodology/linguistic_analysis}

\input{sections/methodology/topic_modeling}

\input{sections/methodology/judgmental_analysis}

%% file: sections/methodology/linguistic_analysis.tex
\subsubsection{Linguistic Analysis}
\label{subsubsec:linguistic_analysis}

\input{tables/linguistic_analysis}

We utilized the TextRank algorithm~\cite{mihalcea-tarau-2004-textrank}, implemented in the PyTextRank\footnote{https://pypi.org/project/pytextrank/} library, to identify key phrases frequently associated with mental health distress. TextRank is a graph-based ranking model that analyzes word relationships within a document, assigning scores to terms based on contextual importance. This method allows us to extract the most informative and representative phrases from mental health discussions.

Through this process, we identified various phrases commonly used in conversations that signal mental disturbance. Example phrases from this set include: \textit{panic attack, my psychiatrist, my mental health, suicidal thoughts, antipsychotics, severe depression, emotional abuse, mental health issues, childhood abuse, childhood trauma, chest pain, manic episodes, hopelessness, trauma therapy, anti-depressant, sexual assault, sleep schedule, social anxiety, and brain damage}. These terms reflect serious emotional distress and serve as important linguistic markers for identifying mental health-related issues in social media discourse.

To further explore the textual differences between posts related to mental conditions and those from the control group, we performed a linguistic analysis focusing on hashtags, URLs, text length, tokenization, and part-of-speech (POS) distribution. The results of this analysis are summarized in Table~\ref{tab:linguistic_analysis}.

First, hashtag usage is relatively low across both groups, with control group posts averaging slightly higher at 0.147 compared to 0.095 in mental health posts. This suggests that mental health discussions on Reddit are less reliant on hashtags, likely due to their more personal and narrative nature.

Additionally, we observed a stark contrast in the presence of URLs within posts. Mental health posts include significantly fewer links, at 3.0\%, compared to 26.1\% in control group posts. This suggests that general discussion topics are more likely to reference external sources, while mental health-related content tends to be more self-reflective and expressive rather than informational.

Another key distinction is post length. Mental health-related posts have an average character count of 1285.28, while control group posts average 852.12 characters. This suggests that users discussing mental health issues tend to write longer, more detailed posts, possibly to express their experiences, emotions, and challenges fully. Correspondingly, the number of tokens per post is also higher for mental health-related posts, with an average of 114.03 tokens compared to 86.59 in control group posts. This supports the intuition that discussions of mental health are typically more elaborate and descriptive.

For part-of-speech (POS) tagging, we used the Penn Treebank~\cite{marcus1993building} tagset, as implemented in the NLTK library. The average counts for various POS tags are shown in Table~\ref{tab:linguistic_analysis}. We observed that mental health-related posts feature a higher frequency of pronouns (36.25 vs. 11.57), verbs (55.52 vs. 25.25), and adjectives (18.68 vs. 11.32), while the frequency of nouns is relatively comparable (46.69 vs. 42.87). The greater use of pronouns in mental health posts indicates a personal, first-person narrative style, as users often discuss their struggles and feelings from their perspective. The increased frequency of verbs and adjectives in these posts suggests that discussions about mental health tend to be more dynamic and emotionally expressive.

These results highlight several important linguistic features that differentiate discussions of mental health from other types of discourse. They indicate that Reddit users discussing mental health tend to write in a more personal, expressive, and emotionally charged manner, often using first-person narration. In contrast, control group posts are shorter, more factual, and more likely to reference external sources.

%% file: tables/linguistic_analysis.tex
\begin{table}[htb]
    \caption{Linguistic Analysis of Mental Disorder and Control Group Posts}
    \centering
    \begin{tabular}{|c||c|c|}
        \hline
        \textbf{Linguistic Metrics} & \textbf{Mental Disorder} & \textbf{Control Group} \\
        \hline
        Avg. hashtag usage & 0.095 & 0.147 \\
        \hline
        Avg. posts containing URLs & 0.030 & 0.261 \\
        \hline
        Avg. character count per post & 1285.28 & 852.12 \\
        \hline
        Avg. token count per post & 114.03 & 86.59 \\
        \hline
        Avg. nouns & 46.69 & 42.87 \\
        \hline
        Avg. pronouns & 36.25 & 11.57 \\
        \hline
        Avg. verbs & 55.52 & 25.25 \\
        \hline
        Avg. adjectives & 18.68 & 11.32 \\
        \hline
    \end{tabular}
    \label{tab:linguistic_analysis}
\end{table}

%% file: sections/methodology/topic_modeling.tex
\subsubsection{Topic Modeling}
\label{subsubsec:topic_modeling}

We applied topic modeling to better understand how users express their thoughts about different mental health conditions. This technique provides valuable insight into the evolving language usage and the degree of emotional distress reflected in posts, further validating our classification of content related to mental disorders. We utilized Latent Dirichlet Allocation (LDA), a popular topic modeling technique that applies unsupervised learning to discover underlying thematic patterns in text documents~\cite{jelodar2019latent, Carron-Arthur2016}.

\begin{figure}
    \centering
    \includegraphics[width=0.8\linewidth]{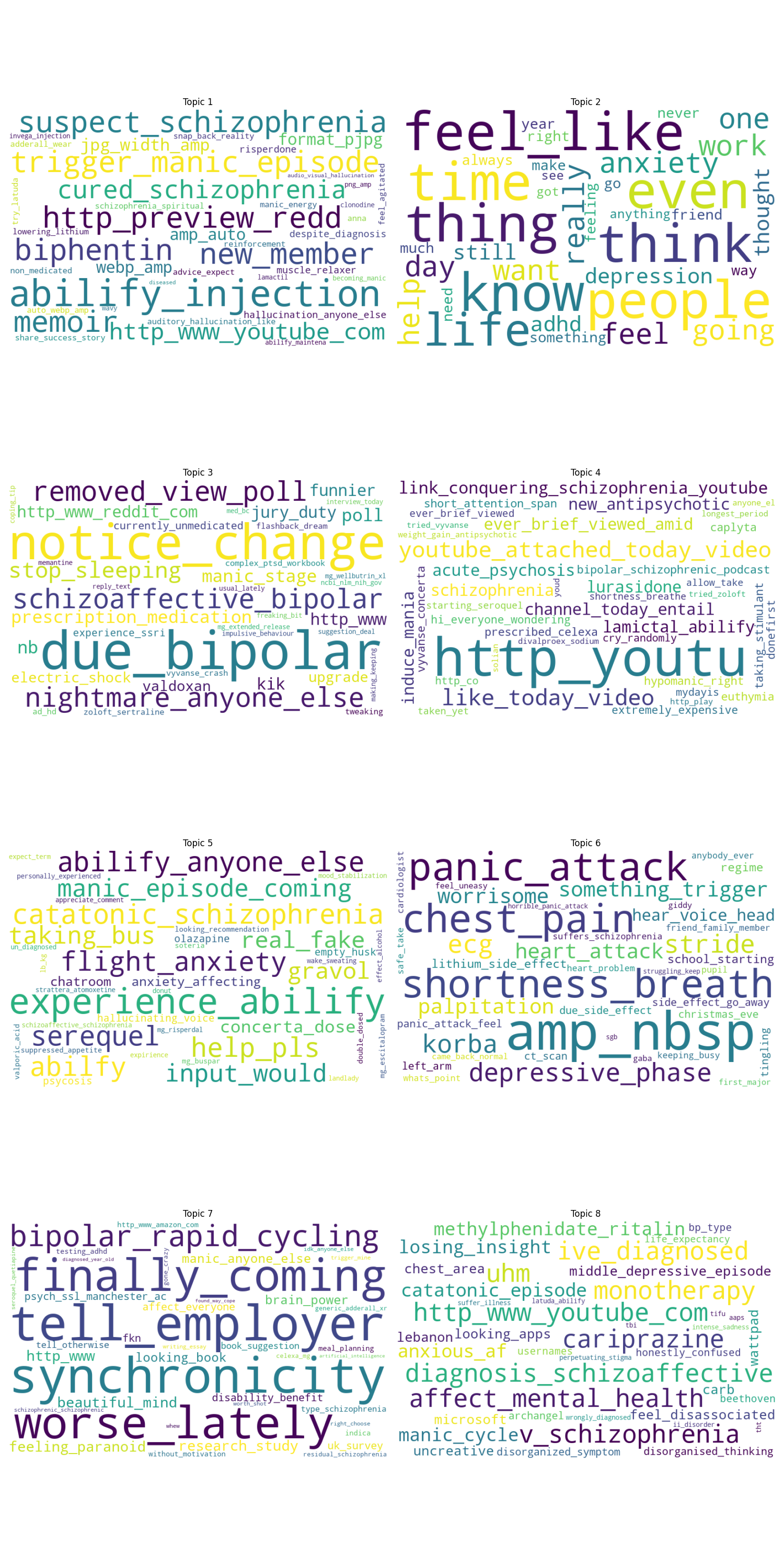}
    \caption{Results of Topic Modeling with eight Topics}
    \label{fig:topic_modeling}
\end{figure}

In our study, we trained an LDA model on user posts to identify prevalent discussion topics in mental health-related subreddits. Text preprocessing involved extracting bigram and trigram phrases using the GENSIM\footnote{https://pypi.org/project/gensim/} library from each post's title and content. These phrases were then transformed into a Term Frequency-Inverse Document Frequency (TF-IDF) representation, used as input for the scikit-learn\footnote{https://scikit-learn.org/} LDA implementation. We experimented with different topic numbers ranging from 1 to 30. Figure~\ref{fig:topic_modeling} presents the results of our LDA topic model, highlighting a sample of eight topics. Keywords indicative of distress, such as \textit{chest pain, panic attack, suspect schizophrenia, anxiety, nightmare, bipolar,  depression, manic episode,  depressive phase, shortness breath, hear voice, help pls, losing insight, short attention span}, frequently appear among the top terms in the extracted topics. These terms further emphasize the prevalence of discussions about psychological distress and crises within these mental health communities. These findings support the relevance of our classification approach and confirm that posts in these subreddits often address serious struggles with mental health.

%% file: sections/methodology/judgmental_analysis.tex
\subsubsection{Judgmental Analysis}
\label{subsubsec:judgmental_analysis}

We enhanced the topic modeling described above with evaluations based on human judgment, in line with established practices in the research literature~\cite{hasan2021survey, Landis1977}. We sampled 1,000 posts stratified by annotated groups from our dataset, which were manually labeled by two annotators. We then calculated the level of agreement between the original and manual annotations for each annotator by applying four metrics: 1) percent agreement, 2) Cohen's $\kappa$~\cite{Cohen1960}, 3) Scott’s $\pi$~\cite{scott1955}, and 4) Krippendorff’s $\alpha$~\cite{krippendorff2004}. A sufficiently high $\kappa$ value indicates that the labeling scheme is consistent and could be generalized to the entire dataset~\cite{Landis1977}. This assertion holds for the other metrics as well, with scores ranging between 0 and 1. This step strengthens the reliability of our annotation methodology. 

Before labeling the dataset, we developed a clear set of guidelines for classifying posts into Mental Disorder and Control Group categories. These guidelines were designed to promote consistent categorization and minimize bias, ensuring the validity of our dataset.

\begin{enumerate} \item \textbf{Mental Disorder Category:} A post is annotated as belonging to the mental disorder category if:
    \begin{itemize}
        \item The post explicitly discusses personal struggles related to mental health conditions such as ADHD, anxiety, bipolar disorder, PTSD, depression, schizophrenia, or BPD.
        
        \item The author describes experiences of emotional distress, including but not limited to panic attacks, depressive episodes, or an active search for support related to a diagnosed or suspected mental health condition.
        
        \item The language used indicates severe emotional distress, such as expressions of hopelessness, extreme anxiety about social interactions, self-harm or suicidal tendencies, or an inability to cope with daily life.
        
        \item The post contains rhetorical expressions that describe emotional experiences, such as \textit{I feel empty}, \textit{I no longer have the strength to carry on}, or \textit{My anxiety is ruining my life}.
    \end{itemize}

\item \textbf{Control Category:} A post is categorized as belonging to the control group if:
    
    \begin{itemize}
        \item There is no mention of personal struggles related to mental health issues.
        
        \item The content discusses broader topics, such as politics, sports, travel, hobbies, professions, or entertainment, with no direct connection to mental health.
        
        \item The post is primarily informative, opinion-based, or part of a casual discussion, rather than a personal account of emotional distress.
        
        \item The author describes hardships or setbacks in a problem-oriented manner, without displaying intense emotional strain. For example, \textit{I had a bad day at work today, but I'll get through it}.
    \end{itemize}
\end{enumerate}

\input{tables/annotator_agreement}

Table~\ref{tab:judgmental_score} reports the measurement of every agreement metric mentioned above for each annotator. We perceive that the agreement metrics strongly point to the reliability of the annotation process. The scores for percent agreement stand at 98.1\% and 97.1\%, indicating a high level of consistency among the annotators. Similarly, Cohen's $\kappa$ values, 0.962 and 0.942, confirm the near-perfect agreement, as by Landis et al., $\kappa > 0.8$ signifies almost perfect concordance~\cite{Landis1977}. Additionally, Scott's $\pi$ and Krippendorff's $\alpha$ tend toward 1 and present results that are coherent with them, strengthening the robustness and reproducibility of the labeling process. These results validate the quality of the annotations provided in the dataset to be of very minimal subjective bias, hence increasing the credibility of further analyses.

%% file: tables/annotator_agreement.tex
\begin{table}[htb]
    \centering
    \caption{Inter-annotator Agreement Metrics}
    \begin{tabular}{c|cc}
        \hline
        \textbf{Metric} & \textbf{Annotator 1} & \textbf{Annotator 2} \\
        \hline
        Percent Agreement & 98.1\% & 97.1\% \\
        Scott's $\pi$ & 0.962 & 0.942 \\
        Cohen's $\kappa$ & 0.962 & 0.942 \\
        Krippendorff's $\alpha$ & 0.962 & 0.942 \\
        \hline
    \end{tabular}
    \label{tab:judgmental_score}
\end{table}

%% file: sections/methodology/classification_models.tex
\subsection{Classification Models}
\label{subsec:classificatiton_models}

We utilized several state-of-the-art transformer-based text classification models to extract meaningful information from raw subreddit posts. The classifiers we trained are as follows:

\begin{itemize}
    \item \textbf{BERT:} BERT (Bidirectional Encoder Representations from Transformers) is a widely used model for natural language processing tasks. It learns the contextual relationships between words in a text by considering both left and right contexts, enabling it to understand the meaning of a word based on its surrounding words~\cite{devlin2019bert}.

    \item \textbf{RoBERTa:} RoBERTa is a variant of BERT optimized by training with larger batch sizes, more data, longer sequences, and removing the next sentence prediction objective. It also fine-tunes key hyperparameters to enhance BERT’s pre-training procedure~\cite{Liu2019RoBERTaAR}.

    \item \textbf{DistilBERT:} DistilBERT is a smaller, faster version of BERT, which retains most of BERT's accuracy while being computationally more efficient. It achieves this through knowledge distillation, where the knowledge of the larger BERT model is transferred to a smaller model~\cite{Sanh2019DistilBERTAD}.

    \item \textbf{ALBERT:} ALBERT (A Lite BERT) is a lighter version of BERT designed to reduce model size and complexity. It achieves this through factorizing embedding parameters and sharing parameters across layers, which allows it to retain performance while being more computationally efficient~\cite{Lan2019ALBERTAL}.

    \item \textbf{ELECTRA:} ELECTRA introduces a new pre-training objective that focuses on training the model on actual tokens versus those replaced by a generator. This approach significantly improves sample efficiency compared to the Masked Language Modeling (MLM) used in BERT~\cite{Clark2020ELECTRA}.

    \item \textbf{LSTM:} We also applied several Long Short-Term Memory (LSTM) models, both with and without attention mechanisms, so the model can learn the relevant context for a long period using a word's left and right contexts. Levels of attention allow for learning pertinent words from a sentence. In addition, we tried model training both with and without bi-directional features. Each model has 128 LSTM units inside, with a dropout rate of 0.2 and a recurrent dropout rate of 0.2. Another dropout layer of 0.2 was added after the attention layer upon creation. Two more fully connected layers were employed at the back end of the models with 256 and 2 units, respectively. On top of implementing LSTM models, we experimented with three context-aware pre-trained embeddings provided by BERT, GloVe, and Word2Vec for the word embedding layer.
\end{itemize}

%% file: sections/results.tex
\section{Experiments and Results}

In all of our experiments, we utilized 5-fold stratified cross-validation to ensure a rigorous and unbiased model evaluation. The primary dataset of 120,000 posts was strategically divided using stratified sampling: 80\% was allocated for training and validation, while the remaining 20\% was reserved as a \textbf{hold-out} test set. This process creates five independent dataset splits, resulting in five distinct versions of each model being trained. This cross-validation approach enables a comprehensive assessment of model performance across diverse training configurations while enhancing the robustness of our evaluation methodology.

Hyperparameter optimization was conducted using Ray Tune\footnote{https://www.ray.io/}, a powerful tool that automates grid searches across possible hyperparameter values to identify the optimal configuration. After an extensive search, generally speaking, the best-performing models shared the same learning rate of $10^{-6}$ and weight decay of $10^{-2}$, which yielded high performance across most of the tested architectures. We employed the AdamW optimizer~\cite{loshchilov2018decoupled} with a learning rate scheduler adapted to the patience of five epochs. This would ensure that the learning rate is only adjusted if performance plateaus, avoiding a potential unnecessary oscillation during training.

All models were implemented using PyTorch and HuggingFace, leveraging their comprehensive natural language processing tools. The computational infrastructure comprised two NVIDIA Ampere A100 GPUs, each featuring 6912 CUDA cores and 40GB of RAM, complemented by a system memory of 512 GB. This infrastructure is quite capable of efficiently training complex transformer-based architectures.

In classifying the hold-out test set, the final label for each test sample was decided by aggregating predictions from the five cross-validation folds. The most frequent class label across these five models was assigned as the final prediction. This ensemble-based approach helps mitigate individual model biases and improves overall prediction stability. Likewise, when evaluating the models on the \textbf{external} test dataset (described in subsection \ref{subsection-DataProcessing}), we applied the same majority voting approach to ensure consistency in our comparative analysis.

Table~\ref{tab:model_performance} compares models on the performance of the hold-out and external test datasets based on the evaluation metrics: accuracy, F1 score, precision, and recall. For completeness, we additionally incorporate an approximation of the training time to produce the best-performing model for each architecture to offer an idea about the computational complexity associated with the different approaches. These analyses provide insight into how well different architectures perform and to what extent their performances generalize to unseen data.

\input{tables/model_performance}

\input{sections/research_questions/RQ1}
\input{sections/research_questions/RQ2}
\input{sections/research_questions/RQ3}

%% file: tables/model_performance.tex
\begin{table*}[htb]
    \centering
    \begin{threeparttable}
        \caption{Model Performance Summary Based on Accuracy, F1 Score, Precision, and Recall on Hold-Out and External Test Sets
        }
        \label{tab:model_performance}
        
        \begin{tabular}{|c|c||c|c|c|c||c|c|c|c||c|}
        \hline
        \textbf{Embedding} & \textbf{Model} & \multicolumn{4}{|c|}{\textbf{Hold-Out Test Dataset}} & \multicolumn{4}{|c|}{\textbf{External Test Dataset}} & \textbf{Time} \\
        \cline{3-10}
        & & \textbf{Accuracy} & \textbf{F1 Score} & \textbf{Precision} & \textbf{Recall} & \textbf{Accuracy} & \textbf{F1 Score} & \textbf{Precision} & \textbf{Recall} & \textbf{(hrs)\tnote{*}} \\
        \hline
        Roberta & Roberta & \textbf{99.54} & \textbf{99.54} & \textbf{99.50} & \textbf{99.58} & \textbf{96.05} & \textbf{95.96} & 98.32 & 93.71 & 1.29 \\
        DistilBERT & DistilBERT & 99.40 & 99.40 & 99.29 & 99.52 & 95.13 & 94.97 & 98.12 & 92.02 & 1.02 \\
        ALBERT & ALBERT & 99.41 & 99.41 & 99.42 & 99.40 & 95.70 & 95.60 & 97.99 & 93.32 & 1.64 \\
        BERT & BERT & 99.49 & 99.49 & 99.42 & 99.56 & 95.22 & 95.06 & 98.43 & 91.91 & 2.02 \\
        ELECTRA & ELECTRA & 99.50 & 99.50 & 99.48 & 99.51 & 95.52 & 95.36 & \textbf{98.74} & 92.21 & 1.91 \\
        \hline
        BERT & BiLSTM + Attention & 98.96 & 98.96 & 98.86 & 99.06 & 94.91 & 94.76 & 97.77 & 91.93 & 0.29 \\
        BERT & BiLSTM & 98.54 & 98.54 & 98.16 & 98.93 & 95.06 & 95.11 & 94.11 & \textbf{96.13} & 0.42 \\
        BERT & LSTM + Attention & 98.99 & 98.99 & 98.85 & 99.13 & 94.51 & 94.31 & 97.77 & 91.09 & \textbf{0.27} \\
        BERT & LSTM & 98.81 & 98.81 & 98.40 & 99.23 & 95.14 & 95.11 & 95.57 & 94.66 & 0.60 \\
        \hline
        GloVe & BiLSTM + Attention & 81.27 & 80.90 & 82.52 & 79.34 & 79.06 & 80.39 & 75.58 & 85.86 & 12.38 \\
        GloVe & BiLSTM & 50.0 & 0 & 0 & 0 & 50.0 & 0 & 0 & 0 & - \\
        GloVe & LSTM + Attention & 81.47 & 81.61 & 81.00 & 82.23 & 78.60 & 79.95 & 75.23 & 85.30 & 6.74 \\
        GloVe & LSTM & 50.0 & 0 & 0 & 0 & 50.0 & 0 & 0 & 0 & - \\
        \hline
        Word2Vec & BiLSTM + Attention & 81.59 & 81.67 & 81.32 & 82.02 & 77.92 & 79.42 & 74.37 & 85.20 & 4.23 \\
        Word2Vec & BiLSTM & 50.0 & 0 & 0 & 0 & 50.0 & 0 & 0 & 0 & - \\
        Word2Vec & LSTM + Attention & 82.35 & 83.26 & 79.19 & 87.78 & 81.29 & 82.33 & 77.97 & 87.21 & 3.36 \\
        Word2Vec & LSTM & 50.0 & 0 & 0 & 0 & 50.0 & 0 & 0 & 0 & - \\
        \hline
        \end{tabular}
        \begin{tablenotes}[flushleft]
            \centering
                \item[*] Approx. training time to achieve a best-performing model; this may vary depending on hardware specifications.
        \end{tablenotes}
    \end{threeparttable}

\end{table*}

%% file: sections/research_questions/RQ1.tex
\subsection{How competitive are transformer-based models in identifying posts of mental illness?}

Transformer-based models perform remarkably well on mental disorder-related post-detection tasks. RoBERTa shows the best performance, with 99.54\% accuracy and F1 score on the hold-out test dataset and a robust 96.05\% accuracy and 95.96\% F1 score on the external test dataset. These metrics indicate the model's good generalization capability to unseen data with well-balanced precision and recall. The transformer architecture, intrinsically combining self-attention mechanisms with deep contextual embeddings, gives the best results on tasks where nuances of texts and context sensitivity are epitomized.

Models such as ELECTRA and BERT perform very closely to RoBERTa, with scores higher than 99\% on hold-out datasets and above 95\% on external datasets. This is representative of the competitiveness achieved by transformer architectures. The drop in the score in the case of external datasets hints at the challenging task of adapting to different linguistic variations and datasets. This emphasizes the need for fine-tuning and data augmentation for robustness.

%% file: sections/research_questions/RQ2.tex
\subsection{How does integrating different word embeddings, such as BERT, GloVe, and Word2Vec, affect the performance of LSTM-based models, and what contribution does the attention mechanism make toward their effectiveness?}

The choice of word embedding significantly impacts the learning capability of LSTM-based models for identifying mental health-related text. Our results reflect a very crucial disparity between transformer-based embeddings and traditional static embeddings regarding their capability to support LSTM models.

Transformer-based embeddings, such as BERT, are rich in contextual information, enabling models like LSTMs to achieve strong classification results even without additional attention mechanisms. For example, a BERT-BiLSTM model without an attention layer in the LSTM component achieves an F1 score of 95.11\% on the external test dataset. This highlights the significant contribution of the BERT embeddings to the model's performance, as their high-quality representations effectively support the learning process.

Static embeddings (e.g., GloVe and Word2Vec), however, face significant limitations when used with preprocessed text. In our experiments, preprocessing steps such as tokenization and removal of unnecessary words and characters resulted in LSTM models failing to train effectively, yielding an F1 score of 0.0. This is likely because static embeddings cannot dynamically adapt word meanings based on context, which is particularly problematic for long texts. Additionally, preprocessing may inadvertently remove critical semantic and syntactic relationships, making it difficult for LSTMs to model meaningful dependencies in long sequences.

Even when using raw text (i.e., unprocessed input), LSTM models with static embeddings only achieve successful training when augmented with an attention mechanism. Without attention, these models struggle to capture long-range dependencies due to issues like vanishing gradients, leading to poor generalization. The attention mechanism addresses this by enabling the model to focus on the most relevant words in the sequence, effectively compensating for the static nature of the embeddings. In our experiments, a GloVe-BiLSTM model with attention achieved an F1 score of 80.39\% on the external test set. While this is a decent result, it still falls short of the performance achieved by transformer-based models like BERT, highlighting the advantages of contextualized embeddings.

Our results highlight that while BERT-based LSTMs do not outperform standalone transformer models, they offer a practical and efficient alternative for resource-constrained environments. These findings underscore the importance of embedding selection and architecture modifications, such as integrating pre-trained contextualized embeddings and optimizing LSTM layers, as key strategies for further improving the performance of LSTM-based models in mental health text classification tasks.

%% file: sections/research_questions/RQ3.tex
\subsection{How robust are the models in terms of computational complexity?}

Computational complexity represents a critical evaluation metric for machine learning models, particularly in sensitive domains like mental health diagnostics. Transformer-based models, while demonstrating exceptional accuracy, come with substantial computational overhead. RoBERTa and BERT, for instance, require approximately 1.29–2.02 hours of training, highlighting the significant computational resources needed for these advanced architectures. Although this computational intensity is justifiable given the critical nature of mental health applications, it poses challenges in resource-limited settings.

LSTMs with transformer embeddings, e.g., BERT-LSTM with attention mechanism, emerge as a compelling compromise. These hybrid models achieve a state-of-the-art 94.31\% F1-score on the external dataset while maintaining a more modest training time of 0.27-0.60 hours. This balance between performance and efficiency makes them particularly attractive for practical deployments with constrained computational resources.
In contrast, traditional embedding approaches like GloVe and Word2Vec-based LSTM models demonstrate less favorable performance characteristics. The attention-enhanced GloVe BiLSTM, for example, required an extensive 12.38 hours of training while only achieving an F1 score of 80.39\% on the external dataset. These findings underscore the limitations of traditional embedding techniques in nuanced text classification tasks, especially those involving complex linguistic patterns like mental health detection.

%% file: sections/conclusion.tex
\section{Conclusion and Future Work}

This paper presented a detailed comparative analysis of transformer-based models and LSTM variants for detecting mental health distress in a large dataset curated from Reddit. Based on our findings, the key conclusions are as follows: 

\begin{enumerate}
    \item Transformer models outperform other architectures, with RoBERTa achieving the highest F1-scores of 99.54\% on the hold-out test dataset and 95.96\% on the external test dataset. However, all transformer models perform competitively, with accuracy and F1-score differences of less than 1\% compared to RoBERTa on both test datasets. This demonstrates the robustness of transformer models for mental health classification tasks.

    \item LSTM models with BERT embeddings perform on par with transformer-based models, achieving F1-scores above 94\% on the external dataset. This near-equivalence in performance highlights the effectiveness of BERT embeddings in LSTM architectures. Additionally, BERT-based embeddings significantly outperform traditional embedding techniques, such as GloVe and Word2Vec, in LSTM models.

    \item While the experimental models exhibit strong generalization capabilities, their performance on external datasets shows slight degradation compared to the hold-out datasets. This suggests a need for more diverse and representative training data to improve model robustness across different datasets.

    \item Although computational efficiency has been improved with models like DistilBERT, training transformer-based architectures still requires significantly more computational resources than LSTM-based models with BERT embeddings. This makes the combination of transformer embeddings and LSTM variants a highly viable alternative for resource-constrained environments.
\end{enumerate}

Future work will address this study's limitations through several key directions. First, we will explore domain-adaptive pretraining methods to enhance the generalization performance of transformer models across diverse social media platforms. Second, we aim to introduce multi-modal approaches by integrating text with additional signals, such as user behavior and sentiment analysis, to improve the robustness of classification. Additionally, we plan to investigate few-shot and zero-shot learning techniques to reduce reliance on large annotated datasets. Finally, we will enhance the interpretability of these models using explainable AI techniques, ensuring transparency and usability in real-world mental health applications.

This research contributes to advancing deep learning methodologies for mental disorder detection, paving the way for developing scalable, real-time monitoring tools. Such tools possess the potential to enable early intervention and improve mental health outcomes on a global scale.